\documentclass[10pt,twocolumn,letterpaper]{article}

\usepackage{iccv}
\usepackage{times}
\usepackage{epsfig}
\usepackage{graphicx}
\usepackage{amsmath, bm}
\usepackage{amssymb}
\usepackage{enumitem}

\usepackage{tabu}
\usepackage{booktabs}
\usepackage{array}

\usepackage{color}
\usepackage{colortbl}
\usepackage[dvipsnames]{xcolor}

\iccvfinalcopy

\usepackage[pagebackref,breaklinks,colorlinks,filecolor=blue, urlcolor=purple, citecolor=RoyalBlue]{hyperref}
\usepackage{multirow}
\usepackage[capitalize]{cleveref}
\usepackage{xcolor}
\usepackage{booktabs}
\usepackage{bbm}
\usepackage{graphicx}
\usepackage{makecell}
\usepackage[dvipsnames]{xcolor}
\usepackage[accsupp]{axessibility}  %

\definecolor{orange}{rgb}{1,0.5,0}
\definecolor{mgray}{gray}{.9}

\newcommand{\modelname}{LATR}

\newcolumntype{x}{>{\columncolor[HTML]{EFEFEF}[16pt]}r}
\newcolumntype{y}{>{\columncolor[HTML]{EFEFEF}[0pt]}l}

\ificcvfinal\fi

\def\eg{\emph{e.g}\onedot} 
\def\ie{\emph{i.e}\onedot} 
 
 \def\vs{\emph{vs}\onedot}
\renewcommand{\wrt}{\textit{w}.\textit{r}.\textit{t}. }

\crefname{section}{Sec.}{Secs.}
\Crefname{section}{Section}{Sections}
\Crefname{table}{Table}{Tables}
\crefname{table}{Tab.}{Tabs.}

\begin{document}

\title{LATR: 3D Lane Detection from Monocular Images with Transformer}
\author{Yueru Luo\textsuperscript{\rm 1,2} \quad Chaoda Zheng\textsuperscript{\rm 1,2}  \quad Xu Yan\textsuperscript{\rm 1,2}
\quad Tang Kun\textsuperscript{\rm 3} 
\quad Chao Zheng\textsuperscript{\rm 3} \quad \\
\quad Shuguang Cui\textsuperscript{\rm 2,1} \quad Zhen Li\textsuperscript{\rm 2,1}\thanks{Corresponding author.} \\
\textsuperscript{\rm 1} FNii, CUHK-Shenzhen \quad
\textsuperscript{\rm 2} SSE, CUHK-Shenzhen \quad
\textsuperscript{\rm 3} Tencent Map, T Lab\\
{\tt\small \{{222010057@link.},{chaodazheng@link.},{218012048@link.},{lizhen@}\}cuhk.edu.cn}
}

\maketitle
\ificcvfinal\thispagestyle{empty}\fi

\begin{abstract}
   \vspace{-0.1in}

3D lane detection from monocular images is a fundamental yet challenging task in autonomous driving.
Recent advances primarily rely on structural 3D surrogates (\eg, bird's eye view)
built from front-view image features and camera parameters.
However, the depth ambiguity in monocular images inevitably causes misalignment between the constructed surrogate feature map and the original image, posing a great challenge for accurate lane detection.
To address the above issue, we present a novel \textbf{\modelname}~model, an end-to-end 3D lane detector that uses 3D-aware front-view features without transformed view representation.
Specifically, \modelname~detects 3D lanes via cross-attention based on query and key-value pairs, constructed using our lane-aware query generator and dynamic 3D ground positional embedding.
On the one hand, each query is generated based on 2D lane-aware features and adopts a hybrid embedding to enhance lane information.
On the other hand,
3D space information is injected as positional embedding from an iteratively-updated 3D ground plane.
\modelname~outperforms previous state-of-the-art methods on both synthetic Apollo, realistic OpenLane and ONCE-3DLanes datasets by large margins (\eg, \textbf{11.4} gain in terms of F1 score on OpenLane). 
Code will be released at \url{https://github.com/JMoonr/LATR}.

\end{abstract}

\section{Introduction}
\begin{figure}[t]
\vspace{-2mm}
\begin{center}
   \includegraphics[width=1.0\linewidth]{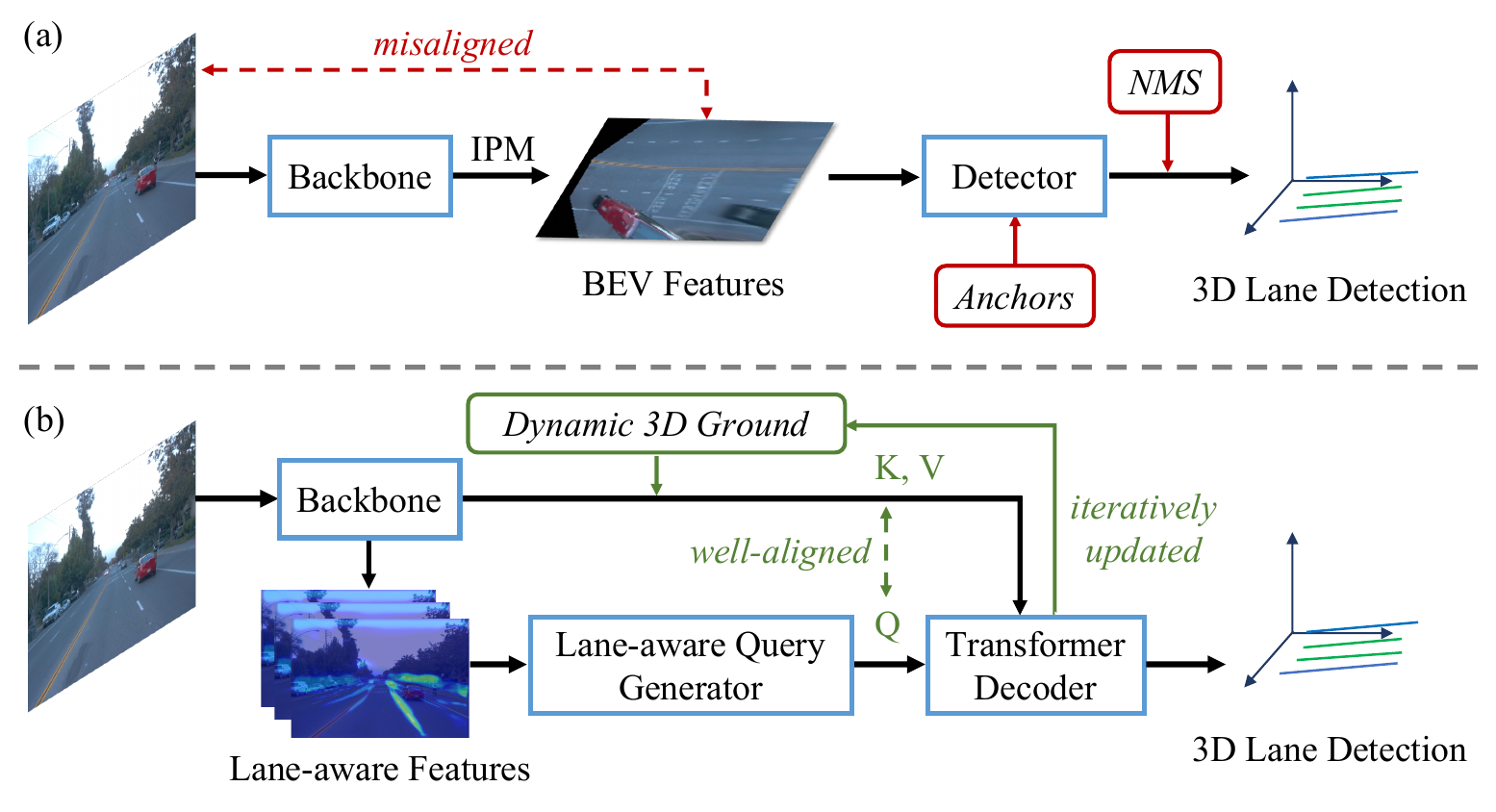}
\end{center}
\vspace{-2mm}
   \caption{(a)~Previous methods mainly utilize camera parameters and inverse perspective mapping (IPM) to transform the features into a surrogate space (\eg, BEV), and further perform 3D lane detection through anchors and non-maximum suppression (NMS). 
   (b)~We propose the novel \modelname, an anchor-free and NMS-free Transformer architecture, to perform 3D lane detection right on the front view. Using lane-aware queries and dynamic 3D ground positional embedding, our model produces well-aligned 3D features and achieves superior 3D lane detection.}
\label{fig:long}
\label{fig:onecol}
\end{figure}

3D Lane Detection is critical for various applications in autonomous driving, such as trajectory planning and lane keeping~\cite{williams2022trajectory}.
Despite the remarkable progress of LiDAR-based methods in other 3D perception tasks~\cite{zhou2018voxelnet,lang2019pointpillars}, recent advances in 3D lane detection prefer using a monocular camera since it owns desirable advantages compared to LiDARs.
Apart from the low deployment cost, cameras offer a longer perception range compared to other sensors and can produce high-resolution images with rich textures, which are crucial for detecting slim and long-span lanes.

Due to the lack of depth information, detecting 3D lanes from monocular images is challenging.
A straightforward solution is to reconstruct the 3D lane layout based on 2D segmentation results and per-pixel depth estimation, as proposed in SALAD~\cite{yan2022once}. However, this method requires high-quality depth data for training and heavily relies on the precision of the estimated depth.
Alternatively, CurveFormer~\cite{bai2022curveformer} employs polynomials to model the 3D lane from the front view. While it avoids indefinite view transformation, the polynomial form adopted in their design restricts the flexibility of capturing diverse lane shapes.
In contrast, current mainstream methods favor the utilization of 3D surrogate representations~\cite{garnett20193d,efrat20203d,guo2020gen,liu2021end,chen2022persformer,li2022reconstruct}. These surrogate representations are constructed based on front-view image features and camera parameters, with no reliance on depth information.
Since lanes inherently reside on the road, most of these methods build the 3D surrogate by projecting the image features into a bird's eye view (BEV) via inverse perspective mapping (IPM)~\cite{mallot1991inverse}.
However, IPM is strictly based on the flat ground assumption, thereby introducing misalignment between the 3D surrogate and the original image in many real driving scenarios (\eg, uphill/downhill and bumps). 
This misalignment, entangled with distortions, inevitably hinders the accurate estimation of the road structure and endangers driving safety.
Despite attempts~\cite{chen2022persformer} made to relieve this issue by 
introducing deformable attention~\cite{zhu2020deformable}, the problem of misalignment remains unresolved.

Based on the above observation, we aim to improve 3D lane detection by {\textit{directly locating 3D lanes from the front view without any intermediate 3D surrogates through lane-aware queries}}.
Inspired by the 2D object detector DETR~\cite{carion2020end}, we streamline lane detection as an end-to-end set prediction problem, forming \textbf{LA}ne \textbf{}detection \textbf{TR}ansformer (\textbf{\modelname}).~\modelname~detects 3D lanes from front view images using \textit{lane-aware queries} and \textit{dynamic 3D ground positional embedding}.
We devise a lane representation scheme to describe lane queries, better capturing the properties of 3D lanes. 
Besides, we utilize lane-aware features to offer queries rich semantic and spatial priors. 
Since pure front-view features lack awareness of 3D space, we inject 3D positional information from a hypothetical 3D ground into the front-view features.
This hypothetical ground, initialized as a horizontal grid, undergoes iterative optimization to fit the ground truth road.
Finally, the lane-aware queries interact with 3D-aware features through a transformer decoder, followed by MLPs to produce 3D lane predictions.

Our main contributions are the following:
\begin{itemize}[leftmargin=*]
		\setlength{\itemsep}{0pt}
		\setlength{\parsep}{0pt}
		\setlength{\parskip}{0pt}
    \item We propose \textbf{\modelname},
    an end-to-end 3D lane detection framework based on Transformer.
    By directly detecting 3D lanes from the front view, without using any 3D surrogate representations,~\modelname~offers efficiency and avoids feature misalignment present in prior methods.
    \item  We introduce a lane-aware query generator that uses dynamically extracted lane-aware features to initialize query embeddings. Moreover, a dynamic positional embedding is proposed to bridge 3D space and 2D images, which derives from a constructed 3D ground that is iteratively updated under supervision. 
    \item We conduct thorough experiments on the benchmark datasets of OpenLane, Apollo, and ONCE-3DLanes. Our proposed~\modelname~outperforms previous SoTA methods by large margins (\textbf{+11.4} improvement on OpenLane, \textbf{+4.3} on Apollo, and \textbf{+6.26} on ONCE-3DLanes \wrt F1 score). 
\end{itemize}

\section{Related Work}
\begin{figure*}[t]
    \centering
    \includegraphics[width=0.95\textwidth]{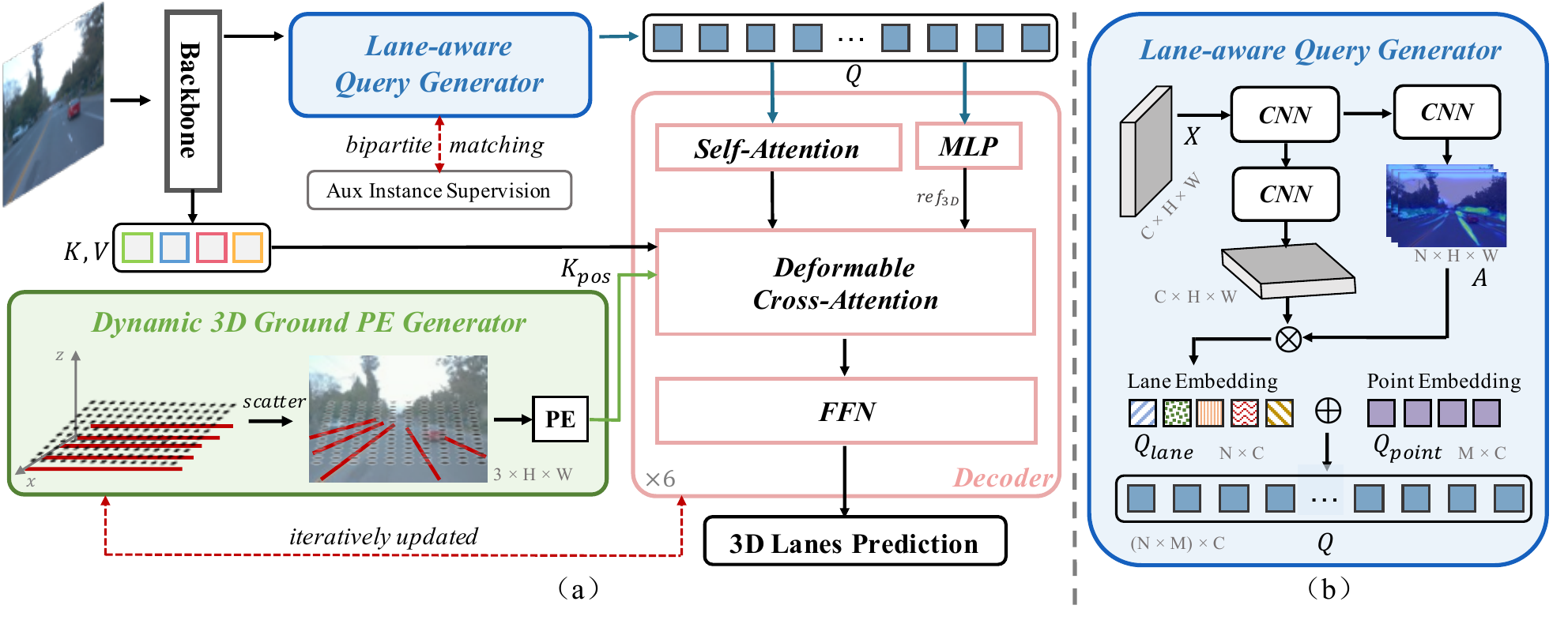}
    \caption{\textbf{The overall architecture.}  LATR is a novel 3D lane detection framework that utilizes a Transformer-based approach, as shown in part (a). Specifically, the front-view image is first processed by the backbone network.
    Subsequently, a Lane-aware Query Generator generates queries that incorporate lane-level and point-level embeddings, as illustrated in (b). In addition, dynamic 3D ground positional embeddings are obtained through iterative refinement of a hypothetical 3D ground plane which is used to capture 3D information.
    }
    \label{fig:arch}
\end{figure*}

\subsection{2D Lane Detection}
\vspace{-1.5mm}
Benefiting from deep learning, 2D lane detection has made significant progress. Primary approaches formulate the problem in four manners. 
\textit{1)~Segmentation-based} methods primarily resort to pixel-wise classification to generate lane masks~\cite{pan2017spatial, neven2018towards, zheng2021resa}, which are further post-processed to form a set of lanes through specific designs. 
\textit{2)~Anchor-based} approaches mainly leverage line-like anchors to regress the offset relative to targets~\cite{li2019line, tabelini2021keep, zheng2022clrnet}. 
To break the limitations of line-shape anchors, ~\cite{jin2022eigenlanes} employs eigenlane space to generate diverse lane candidates.
Besides, row-based anchors are heuristically devised to classify row-wise pixels into lanes~\cite{yoo2020end, qin2020ultra, liu2021condlanenet}.
These row anchors are further extended to hybrid (row and column) anchors in~\cite{qin2022ultra}, attempting to mitigate the localization errors for side lanes.
\textit{3)~Keypoint-based} approaches are proposed to more flexibly model lane positions~\cite{ko2021key, qu2021focus, wang2022keypoint, jin2022eigenlanes}, which first estimate locations of points and then group points into distinct lanes using different schemes. 
\textit{4)~Curve-based} approaches~\cite{van2019end,tabelini2021polylanenet, liu2021end, feng2022rethinking} aim to fit lane curves via various curve formulations. %
Despite these promising advancements in 2D lane detection, a non-negligible gap persists between the 2D results and the requirements of real-world application scenarios, \ie, accurate 3D positions. 
\subsection{3D Lane Detection}\label{sec:rel-3dlane}
To acquire accurate 3D positions in realistic scenarios, 3D lane detection was proposed to drive further related studies. Compared to the 2D lane task, nonetheless, there are fewer studies to explore the model design.
Prevalent methods~\cite{garnett20193d,efrat20203d,guo2020gen,liu2021end,chen2022persformer,li2022reconstruct} attempt to transform the 2D features into a surrogate 3D space based on IPM, operating under the flat road assumption. 
However, this assumption is easily broken when encountering uphill/downhill, which is common in driving scenarios.
Thus, the latent feature representation becomes entangled with unexpected distortions caused by variations in road height, which impairs the model's reliability and jeopardizes traffic safety. 

To tackle this problem, SALAD~\cite{yan2022once} employs front-view image segmentation, utilizing the obtained results to generate 3D lanes with the aid of depth estimation.
However, their training demands dense depth annotation, and their performance hinges on the accuracy of the estimated depth.
GP~\cite{li2022reconstruct} employs 3D lane reconstruction to achieve detection. 
To preserve the 3D lane structure, they explicitly impose geometric constraints on intra-lane and inter-lane relationships under sophisticatedly designed supervision.
M$^2$-3Dlane~\cite{luo2022m}, instead, introduces LiDAR data to facilitate the monocular 3D lane detection. By leveraging LiDAR point clouds, they lift image features into 3D space and further fuse multi-modal features in BEV space.
A concurrent work CurveFormer~\cite{bai2022curveformer} also formulates detection within the perspective space under an end-to-end paradigm.
However, it represents lanes parametrically and forces the model to implicitly learn 3D positions. 
While the polynomial formulation could provide a compact lane structure, it suffers from parameter sensitivity and lacks the flexibility to capture diverse lanes in realistic driving scenarios. 

Different from previous works,~\modelname~relies only on the front view and the monocular image to achieve end-to-end 3D lane detection.
It establishes both intra-lane and inter-lane geometric relationships without imposing hard constraints~\cite {li2022reconstruct}.
Our queries are endowed with lane-aware information from a global perspective, which further integrates with dynamic 3D ground features to predict 3D lanes. 
\subsection{Object Detection with Transformers}\label{sec:rel_cam3d}
DETR~\cite{carion2020end} stands as a pioneer in adopting the end-to-end Transformer architecture for object detection, eliminating the need for heuristic designs in previous CNN-based methods, \eg, anchors, NMS, and instead employing learnable queries to target potential objects.
Recently, a number of subsequent studies
have embraced the end-to-end framework.
To bridge 3D space and 2D image, monocular methods usually require depth estimation for the entire image~\cite{huang2022monodtr,qin2022monoground}, 
or foreground~\cite{zhang2022monodetr}, or object centers~\cite{lu2021geometry}, while multi-view ones involve dense 3D space reconstruction~\cite{li2022bevdepth,li2022bevformer,liu2022bevfusion} or the generation of 3D frustums~\cite{liu2022petr,liu2022petrv2}.

Furthermore, existing query-based object detection efforts~\cite{carion2020end, zhu2020deformable,li2022bevdepth,li2022bevformer,liu2022bevfusion} initialize queries either randomly, or by integrating tailored anchor designs~\cite{wang2021anchor,liu2022dab} for 2D objects. 
For 3D objects, approaches involve generating 3D frustum embeddings~\cite{liu2022petr, liu2022petrv2} to provide localization information.
However, applying these techniques directly to 3D lane detection brings certain limitations.
First, the optimization of ambiguous learnable queries is challenging~\cite{sun2021rethinking} and adversely affects query feature aggregation~\cite{wang2021anchor, liu2022dab}.
Moreover, the static nature of the learned queries hampers their ability to generalize effectively.
Second, due to the significant disparity between compact objects and slender lanes, those tailored designs~\cite{wang2021anchor, liu2022dab} are not suitable (\eg, 
anchor box-based query) or sub-optimal~\cite{zhangmonodetr, liu2022petr} (\eg, frustum in 3D space) for lane detection.

\section{Method}
Given an input image $\mathbf{I} \in \mathbb{R}^{3\times H\times W}$, 3D lane detection aims to predict the 3D position of lanes within it. Lanes are represented by a collection of 3D points, denoted as $\mathbf{Y} = \{ \mathbf{L}_{i} | i \in 1, ..., N\}$, where $N$ is the number of lanes in the image, and $\mathbf{L}_i$ denotes the $i$-th lane. Each lane $\mathbf{L}_i =(\mathbf{P}_i, \mathcal{C}_i)$ consists of a set of points $\mathbf{P}_i=\{(x_i^j,y_i^j,z_i^j)\}_{j=1}^{M}$ that construct the lane, 
where $M$ is a predetermined cardinality of the output point set and $\mathcal{C}_i$ indicates the category.
Normally, $y^{\{*\}}$ is set as a predefined longitudinal coordinate $\mathbf{Y}_{ref} = \{y^i\}_{i=1}^{M}$~\cite{garnett20193d,guo2020gen,chen2022persformer}.
\subsection{Overall Architecture}
The overall architecture of \modelname~is illustrated in~\cref{fig:arch}.
We first extract the feature map $\mathbf{X} \in \mathbb{R}^{C\times H\times W}$ from the input image using a 2D backbone.
Afterward, we generate lane-aware queries $\mathbf{Q} \in \mathbb{R}^{(N \times M) \times C}$ using the Lane-aware Query Generator, with $N$ denoting the number of lanes and each lane being depicted by $M$ points. Here, $(N\times M)$ indicates flattened channels along corresponding dimensions.
Subsequently, the lane-aware queries $\mathbf{Q}$ aggressively interact with the feature map $\mathbf{X}$ via deformable attention~\cite{zhu2020deformable}.
Without constructing any misaligned 3D surrogates, we propose the Dynamic 3D Ground Positional Embedding (PE) Generator to enhance 2D features with 3D awareness during the deformable attention.
Finally, we apply a prediction head on the updated queries to obtain the final lane prediction.
Details of each component will be given in the following subsections.

\subsection{Lane-aware Query Generator} \label{sec:query_gen}
Instead of adopting fixed learned features as queries in previous methods~\cite{carion2020end,li2022bevformer,liu2022petr,zhu2020deformable}, we propose a dynamic scheme to generate lane-aware query embeddings, which provide queries with valuable 2D lane priors derived from image features.
Moreover, to offer queries more flexibility in portraying lanes and implicitly modeling intra-lane and inter-lane relations, we represent the query embedding with diverse granularities, lane-level and point-level.

The lane-level embedding captures the entire structure of each lane, while the point-level embedding gathers local features situated at $y^i \in \mathbf{Y}_{ref}$ as stated above. 
We then concert these two levels as our final query embeddings.
This hybrid embedding scheme encodes queries with discriminative features from different lanes, and enables each query to capture shared patterns at a particular depth by sharing the point-level embedding. 
The intuition here is that points located at the same depth across lanes will undergo a uniform scaling factor during image projection.
Besides, sharing the same road gives rise to shared properties among the lanes in 3D space, \eg, identical elevation.
The inner structure of this module is shown in Fig.~\ref{fig:arch} part (b).

\noindent\textbf{1) Lane-level embedding} encodes features of $N$ lane instances from the image feature map $\mathbf{X} \in \mathbb{R}^{C\times H\times W}$. 
Concretely, we utilize a lane-aware feature aggregator to gather features of distinct lanes, based on a set of instance activation maps (IAMs) $\mathbf{A} \in \mathbb{R}^{N \times (H \times W)}$~\cite{cheng2022sparse}.
The IAMs are dynamically generated and can be formulated as:
\begin{equation}
\setlength{\abovedisplayskip}{6pt}
    \mathbf{A} = \sigma(\mathcal{F}([\mathbf{X}, \mathbf{S}])),\nonumber
\setlength{\belowdisplayskip}{6pt}
\end{equation}
where $\mathcal{F}$ is implemented by several convolution layers, $\sigma$ is a sigmoid function, $[\,,\, ]$ denotes concatenation and $\mathbf{S}$ is a two-channel feature map which represents the 2D spatial localization of pixels~\cite{liu2018intriguing}. 
With IAMs, the lane-level embedding $\mathbf{Q}_{lane} \in \mathbb{R}^{N \times C}$ can be obtained via:
\begin{equation}
\setlength{\abovedisplayskip}{4pt}
    \mathbf{Q}_{lane} = \mathbf{A} \cdot \mathbf{X}^\mathsf{T}.\nonumber
    \setlength{\belowdisplayskip}{4pt}
\end{equation}
During training, we add an auxiliary segmentation head~\cite{cheng2022sparse} on top of $\mathbf{Q}_{lane}$ to predict 2D lane instance masks, which are supervised by projected 3D annotations.
Following \cite{cheng2022sparse}, we assign ground-truth labels to $N$ instance masks using dice-based bipartite matching.
We also use the same matched results to assign labels to our final lane predictions.
Please refer to our supplementary for more details.

\noindent\textbf{2) Point-level embedding} expresses how points relate to each other in a lane.
Instead of deriving it from the image features, we represent it as a set of learnable weights $\mathbf{Q}_{point} \in \mathbb{R}^{M\times C}$,  where each $\bm{q}^i_{point} \in \mathbb{R}^{1\times C}$ corresponds to the embedding of a $y^i$ in the predefined $\mathbf{Y}_{ref}$.
These embeddings will be learned during training.

\noindent\textbf{3) Lane-aware query embedding}  $\mathbf{Q} \in \mathbb{R}^{(N \times M) \times C}$ can be obtained using the following equation: 
\begin{equation}
    \setlength{\abovedisplayskip}{6pt}
         \mathbf{Q} = \mathbf{Q}_{lane} \oplus \mathbf{Q}_{point},\nonumber
    \setlength{\belowdisplayskip}{6pt}
    \end{equation}
where $\oplus$ denotes the broadcasting summation.
The broadcast operation enables the model to distinguish different lanes and group points based on lane-level embeddings.
It is noteworthy that sharing point-level embeddings aids in capturing relative relationships and shared characteristics among points across various lanes, without introducing hard geometric priors, \eg, point-distance constraints and intricate supervision as employed in previous methods~\cite{li2022reconstruct}.

\subsection {Dynamic 3D Ground Positional Embedding} \label{sec:key_pos}
As discussed in~\cref{sec:rel-3dlane}, 
existing methods primarily either utilize a surrogate 3D space to estimate 3D lane positions~\cite{chen2022persformer,guo2020gen,garnett20193d,efrat20203d} or implicitly force the model to learn 3D positions~\cite{bai2022curveformer}. 
Differently, we propose to use the prior that all lanes are located on the ground in the real world and construct a 3D plane to model the ground.
Although there have been several attempts to leverage ground prior to facilitating 3D object detection, some make strong hypotheses like fixed ground~\cite{liu2021ground}, while others introduce extra prediction tasks, \eg, dense depth estimation~\cite{wang2019pseudo,reading2021categorical,qin2022monoground}, horizon line detection, and contact points detection~\cite{yang2022ground}.
However, predicting these extra tasks poses a significant challenge due to their high degree of freedom (DOF), and inaccurate predictions can inevitably undermine performance due to accumulative errors.
In this paper, we tackle this problem by restricting the plane to only two DOF. 
Specifically, we encode the plane as the \textit{key}'s positional embedding in each deformable attention module.
As follows, we will delve into how to generate the \textit{3D ground positional embedding} and update the hypothetical plane to dynamically approach the real ground with only 3D lane annotations.

\noindent\textbf{Positional Embedding Generation.}
We first construct a 3D plane $\mathbf{P} \in \mathbb{R}^{P\times3}$ represented by 3D grids with $P$ points
$\mathbf{P}=\{(x_i, y_i, z_i) | i \in 1, ..., P\}$
and project all points into the 2D image using camera parameters $T$ as:
\begin{equation}\label{equ:project}
 \setlength{\abovedisplayskip}{6pt}
    d * [u, v, 1]^\mathsf{T} = T\cdot [x,y,z,1]^\mathsf{T}.
    \setlength{\belowdisplayskip}{3pt}
\end{equation}
We initialize the grid as a horizontal plane with $z$ empirically set to $-1.5$.
Based on~\cref{equ:project}, we scatter all projected points into a 2D canvas $\mathbf{M}_{p} \in \mathbb{R}^{3 \times H\times W }$, which preserves 3D position for each projected point as:
\begin{equation}\label{equ:scatter}
 \setlength{\abovedisplayskip}{6pt}
    \mathbf{M}_{p}[:, v, u] = (x, y, z),
\setlength{\belowdisplayskip}{3pt}
\end{equation}
where $(u, v)$ and $(x, y, z)$ denotes 2D and 3D coordinates defined in~\cref{equ:project}. For those pixels without projected points, we simply set them to 0s.
Afterward, we obtain the 3D ground positional embedding $\mathbf{PE} \in \mathbb{R}^{C \times H\times W }$ via an MLP.

\noindent\textbf{Dynamic Plane Update.}
In order to dynamically update the plane to approach the real ground, we predict a residual transformation matrix with two DOF (\ie, $\Delta \theta_{x}, \Delta h$) in each decoder layer, using image features and projected canvas,
\begin{equation}
\setlength{\abovedisplayskip}{6pt}
    [\Delta \theta_x, \Delta h] = \text{MLP}(\text{AvgPool}(\mathcal{G}([\mathbf{X}, \mathbf{M}_{p}]))),
    \label{equ:predtz}
\setlength{\belowdisplayskip}{3pt}
\end{equation}
where $[\, , \, ]$
represents concatenation, $\mathcal{G}$ is two convolution layers, $\mathbf{X}$ denotes 2D features from backbone and $\mathbf{M}_{p}$ is the projected canvas that encodes 3D positions from~\cref{equ:scatter}. Further, the transformation matrix can be formulated as:
\begin{equation}
\setlength{\abovedisplayskip}{7pt}
\begin{small}
    D =
    \begin{bmatrix}
        1 & 0 & 0 & 0 \\
        0 & \cos{\Delta \theta_x} & -\sin{\Delta \theta_x} & 0 \\
        0 & \sin{\Delta \theta_x} & \cos{\Delta \theta_x} & \Delta h \\
        0 & 0 & 0 & 1
    \end{bmatrix},
\end{small}
\setlength{\belowdisplayskip}{6pt}
\end{equation}
Then, we iteratively update the plane layer by layer via: 
\begin{equation}
\setlength{\abovedisplayskip}{6pt}
\begin{small}
    \widetilde{\mathbf{P}}_l^\mathsf{T} = D \cdot \widetilde{\mathbf{P}}_{l-1}^\mathsf{T},
\end{small}
\setlength{\belowdisplayskip}{3pt}
\end{equation}
where $\widetilde{\mathbf{P}}$ is the homogeneous representation of $\mathbf{P}$ and $l$ indexes the decoder layer. An algorithm is provided in our Appendix to summarize this procedure.

\begin{figure}[t]
\begin{center}
\includegraphics[width=\linewidth]{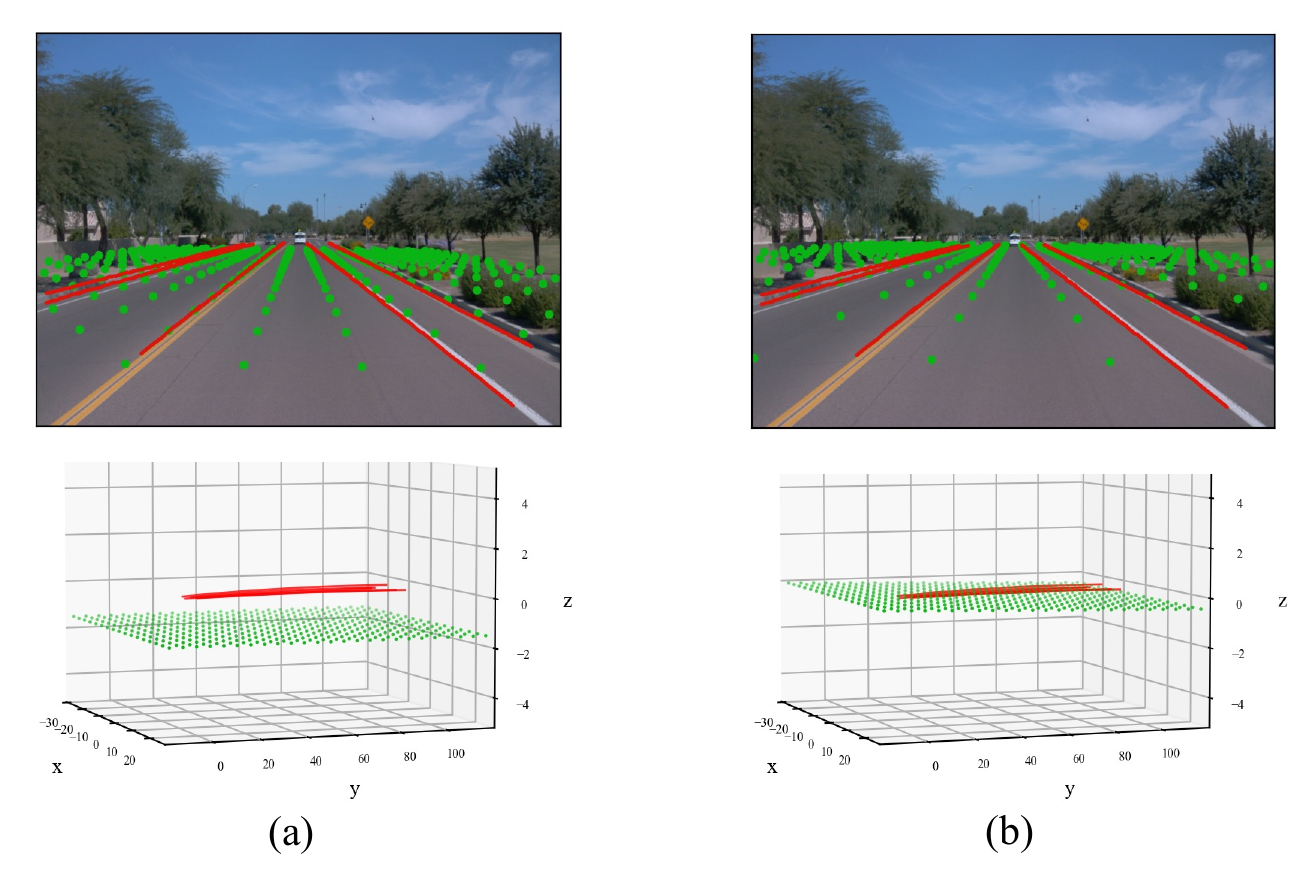}
\end{center}
\vspace{-4mm}
   \caption{\textbf{An illustration of dynamic ground update.} Green points are sampled from our constructed 3D ground plane, and red lines represent ground truth lanes. (a) The left column demonstrates the initial 3D position of the constructed plane (\textcolor{Green}{green}) and the ground truth (\textcolor{red}{red}). (b) The right column shows the updated position, where our constructed ground learned to find the real road position, \ie, they are nearly in the same plane.
   }
\label{fig:plane_upd_case1}
\vspace{-1mm}
\end{figure}

\noindent\textbf{Supervision.}
To supervise the predicted transformation matrix $D$ with two DOF, $\Delta \theta_z$, $\Delta h$.
we project 3D lane annotations onto images using the camera parameters $T$ via: 
$l_{u,v} = T \cdot l_{x,y,z}$,
where $l_{x,y,z}$ is 3D lane annotations and $l_{u,v}$ is the corresponding location on the 2D image.
Similar to~\cref{equ:scatter},  we scatter all projected lanes into $\mathbf{M}_l$.
Thus, we use projected 3D lane annotations to sparsely supervise the projected plane. Given $\mathbb{P}$, which represents the set of all pixels projected from the plane, and $\mathbb{L}$, the set of all
projected 3D lane annotations, our loss can be formulated as:
\begin{equation}
\setlength{\abovedisplayskip}{5pt}
    \mathcal{L}_{plane} = \sum_{u, v \in \mathbb{P}\cap\mathbb{L}}||\mathbf{M}_{p}[:, u, v] - \mathbf{M}_l[:, u, v]||_{2}.\nonumber
\setlength{\belowdisplayskip}{3pt}
\end{equation}
We use $\mathcal{L}_{plane}$ to supervise the update of the constructed 3D plane, allowing it to approach the real ground and obtain accurate 3D ground positional information.
An illustration demonstrating
the efficacy of $\mathcal{L}_{plane}$ is shown in~\cref{fig:plane_upd_case1}.

\subsection{Decoder} \label{sec:decoder}
We build our decoder with $L$ decoder layers following standard Transformer-based methods~\cite{carion2020end,zhu2020deformable}.
In each layer, 
we use the query to predict 3D lane locations $(x,y,z)$ as 3D reference points $\mathit{ref}_{3D}$ and project each 3D point into the 2D image termed as $\mathit{ref}_{2D}$ following~\cref{equ:project}.
Then, we formulate the message exchange process using
the lane-aware query embedding $\mathbf{Q} \in \mathbb{R}^{(N \times M) \times C}$
and the 3D ground positional embedding $\mathbf{PE} \in \mathbb{R}^{C \times (H \times W)}$
as follows:
\begin{equation}
    \setlength{\abovedisplayskip}{6pt}
    \begin{aligned}
        \mathbf{Q}_l = \textrm{DeformAttn}(\mathbf{Q}_{l-1},\mathbf{X}^\mathsf{T} + \mathbf{PE}^\mathsf{T}, \mathbf{X}^\mathsf{T} + \mathbf{PE}^\mathsf{T}, \mathit{ref}_{2D}),\nonumber
    \end{aligned}
    \setlength{\belowdisplayskip}{6pt}
\end{equation}
where $\mathbf{X} \in \mathbb{R}^{C \times (H \times W)}$ represents the extracted image feature map, $l$ is the layer index, and $\textrm{DeformAttn}$ is a standard deformable attention module~\cite{zhu2020deformable}.
As shown in~\cref{fig:arch}, we also estimate the residual transformation matrix of our constructed plane as illustrated in~\cref{sec:key_pos} and iteratively adjust its locations similar to reference points. With this iterative refinement mechanism,~\modelname~can progressively update its knowledge about 3D ground and improve its localizability.

\begin{table*}[t]\label{tab:openlane-main-res}
\centering
  \begin{tabular}{llcllll}
  \toprule
  \multicolumn{1}{c}{} & \multicolumn{1}{c}{} & {{Category } \multirow{-2}{*}{\quad}} & \multicolumn{2}{c}{{X error (m)} $\downarrow$} & \multicolumn{2}{c}{{Z error (m)} $\downarrow$} \\
  \cmidrule{4-5} \cmidrule{6-7} 
  \multicolumn{1}{l}{\multirow{-2}{*}{{Methods} \,}} & \multicolumn{1}{l}{\multirow{-2}{*}{{F1} $\uparrow$}} & {{Accuracy} \multirow{-2}{*}{\, $\uparrow$}} & \multicolumn{1}{l}{\textit{near}} & \multicolumn{1}{l}{\,\textit{far}} & \multicolumn{1}{l}{\textit{near}} & \multicolumn{1}{l}{\,\textit{far}} \\ \hline\hline
  \rule{0pt}{1.1pt} 
  3DLaneNet~\cite{garnett20193d}        & 44.1 &  -  \quad\quad\quad & 0.479 & 0.572 & 0.367 & 0.443 \\
  \rule{0pt}{1.1pt} 
  GenLaneNet~\cite{guo2020gen}          & 32.3 &  -  \quad\quad\quad & 0.593 & 0.494 & 0.140 & 0.195 \\
  \rule{0pt}{1.1pt}
  Cond-IPM                              & 36.6 & - \quad\quad\quad & 0.563 & 1.080 & 0.421 & 0.892 \\
  \rule{0pt}{1.1pt}
  Persformer{$^{\ast}$}~\cite{chen2022persformer}  & \underline{50.5} & \underline{89.5}\quad\quad\quad & \underline{0.319} & \underline{0.325} & \underline{0.112} & \underline{0.141} \\
  \rule{0pt}{1.1pt} 
  CurveFormer~\cite{bai2022curveformer} & \underline{50.5} & -   \quad\quad\quad & 0.340 & 0.772 & 0.207 & 0.651 \\
  
  \rule{0pt}{1.1pt} 
  Persformer-Res50$^\dagger$                  & 53.0 & 89.2\quad\quad\quad & 0.321 & 0.303 & 0.085 & 0.118  \\
  
  \hline 
  \rule{0pt}{1.1pt} 
  \modelname-Lite                & 61.5 & 91.9\quad\quad\quad & 0.225 & 0.249 & 0.073 & 0.106  \\
  \rule{0pt}{1.1pt} 
  \cellcolor{mgray}\modelname %
                                        & \cellcolor{mgray}\textbf{61.9} \footnotesize\color{RoyalBlue}{$\uparrow$\,11.4} %
                                        & \cellcolor{mgray}\textbf{92.0} \footnotesize\color{RoyalBlue}{$\uparrow$\,2.5} %
                                        & \cellcolor{mgray}\textbf{0.219} \footnotesize\color{RoyalBlue}{$\downarrow$\,0.100} %
                                        & \cellcolor{mgray}\textbf{0.259} \footnotesize\color{RoyalBlue}{$\downarrow$\,0.066} %
                                        & \cellcolor{mgray}\textbf{0.075} \footnotesize\color{RoyalBlue}{$\downarrow$\,0.037} %
                                        & \cellcolor{mgray}\textbf{0.104} \footnotesize\color{RoyalBlue}{$\downarrow$\,0.037} \\
  
  \bottomrule
  \end{tabular}
\vspace{1.5mm}
\caption{\textbf{Comparison with other 3D lane detection methods on the OpenLane validation dataset.} $\ast$ denotes the officially released results. $\dagger$ denotes the results reproduced by Persformer with ResNet50 and input images with a shape of 720$\times$960 for a fair comparison. The $\footnotesize\downarrow$ in the head row indicates that lower metric values correspond to better performance, and vice versa. \textbf{Bold} numbers denote the best results and \underline{underline} ones denote the previous best results. Performance gains are highlighted with \textcolor{RoyalBlue}{blue} arrows.}
\label{tab:openlane}
\end{table*}

\begin{table*}[h]
\centering
\scalebox{1.0}{
{
\begin{tabular}{l!{\vrule width 1.2pt}c|c|c|c|c|c|c}
\toprule[1.2pt]
\multirow{2}{*}{Methods} & \multirow{2}{*}{All} & Up \&  & \multirow{2}{*}{Curve} & Extreme & \multirow{2}{*}{Night} & \multirow{2}{*}{Intersection} & Merge  \\
& &  Down & & Weather &  & & \& Split\\
\hline \hline
3DLaneNet~\cite{garnett20193d} & 44.1 & 40.8 & 46.5 & 47.5 & 41.5 & 32.1 & 41.7 \\
GenLaneNet~\cite{guo2020gen} & 32.3 & 25.4 & 33.5 & 28.1 & 18.7 & 21.4 & 31.0 \\
Persformer$^\ast$~\cite{chen2022persformer} & \underline{50.5} & 42.4 & 55.6 & 48.6 & 46.6 & 40.0 & \underline{50.7} \\
CurveFormer~\cite{bai2022curveformer} & \underline{50.5} & \underline{45.2} & \underline{56.6} & \underline{49.7} & \underline{49.1}& \underline{42.9} & 45.4\\
Persformer-Res50$^\dagger$ & 53.0 & 45.7 & 58.5 & 53.8 & 47.5 & 41.3 & 51.5 \\
\hline
\addlinespace[0.3mm]
\modelname-Lite (ours) & 61.5 & 55.2 & 67.9 & \textbf{57.6} & 55.1 & 52.1 & 60.3 \\
\cellcolor{mgray}\modelname~(ours) & 
\cellcolor{mgray}\textbf{61.9}& 
\cellcolor{mgray}\textbf{55.2} & 
\cellcolor{mgray}\textbf{68.2} & 
\cellcolor{mgray} 57.1 & 
\cellcolor{mgray}\textbf{55.4} & 
\cellcolor{mgray}\textbf{52.3} & \cellcolor{mgray}\textbf{61.5} \\
\hline
\textit{Improvement} & \footnotesize\color{RoyalBlue}{{$\uparrow$\,11.4}} &
\footnotesize\color{RoyalBlue}{{$\uparrow$\,10.0}} &
\footnotesize\color{RoyalBlue}{{$\uparrow$\,11.6}} &
\footnotesize\color{RoyalBlue}{{$\uparrow$\,7.4}} &
\footnotesize\color{RoyalBlue}{{$\uparrow$\,6.3}} &
\footnotesize\color{RoyalBlue}{{$\uparrow$\,9.4}} &
\footnotesize\color{RoyalBlue}{{$\uparrow$\,10.8}}\\
\bottomrule
\end{tabular}}}

\vspace{2mm}
\caption{\textbf{Comparison with other 3D lane detection methods on OpenLane dataset under different scenarios.} $\ast$ denotes the latest official results. $\dagger$ denotes the results reproduced by Persformer~\cite{chen2022persformer} with ResNet50~\cite{he2016deep} using the same input size as ours for a fair comparison.
Black bold ones denote the best results.
\textcolor{RoyalBlue}{Blue} arrows denote the performance gain.
\textbf{\modelname-Lite} is the lite version of our~\modelname, which comprises only two decoder layers, while it surpasses all previous SoTA methods and is on par with LATR. \textbf{\textit{improvement}} is calculated by comparing~\modelname~and the previous best results. 
}
\label{tab:openlane-cases}
\vspace{-4mm}
\end{table*}

\subsection{Prediction Head and Losses} \label{sec:head_loss}
\noindent\textbf{Prediction Head.} We apply a prediction head on top of the query
to generate our final predictions.
For the 3D positions estimation, we use an MLP
which can be formulated as:
\begin{equation}
    \setlength{\abovedisplayskip}{6pt}
        [\boldsymbol{\Delta x}, \boldsymbol{\Delta z}, \boldsymbol{v}] = \text{MLP}_{reg}(\mathbf{Q}),\nonumber
    \setlength{\belowdisplayskip}{3pt}
\end{equation}
where $\boldsymbol{\Delta x}, \boldsymbol{\Delta z} \in \mathbb{R}^{N \times M \times 1}$ represents the offsets \textit{w.r.t.} the corresponding reference points (\cref{sec:decoder}) in last decoder layer. 
And $\boldsymbol{v} \in \mathbb{R}^{N \times M \times 1}$ denotes the visibility of each predicted lane point, which indicates whether the projected point is valid in the image.
Together with the predefined longitudinal coordinate $\mathbf{Y}_{ref} \in \mathbb{R}^{M \times 1}$, we obtain $N$ point sets as $\mathit{ref}_{3D}$.
For the lane category, we adopt max-pooling
along the point dimension followed by a per-instance MLP, formulated as:
\begin{equation}
\setlength{\abovedisplayskip}{4pt}
\mathbf{\mathcal{C}} = \text{MLP}_{cls}(\text{MaxPool}(\mathbf{Q})),\nonumber
\setlength{\belowdisplayskip}{4pt}
\end{equation}
where $\mathbf{\mathcal{C}} \in \mathbb{R}^{N \times K}$ denotes the class-logits and $K$ is the number of possible classes.
Lanes classified as ``background" will be discarded in our final predictions.
Here we apply the same bipartite matching results as our auxiliary segmentation in~\cref{sec:query_gen} for 3D lane label assignment. This strategy ensures consistent 2D segmentation and 3D lane detection supervision for each query.

\noindent\textbf{Overall Loss.}
Given matched ground truth labels for the lane queries, we calculate the corresponding loss for each matched pair.
Concretely, our total loss consists of three parts: the instance segmentation auxiliary loss $\mathcal{L}_{seg}$ (\cref{sec:query_gen}), the 3D ground-aware plane update loss $\mathcal{L}_{plane}$ (\cref{sec:key_pos}), and the 3D lane prediction loss $\mathcal{L}_{lane}$. Formally, we have:
\begin{equation}
\setlength{\abovedisplayskip}{6pt}
\begin{aligned}
    \mathcal{L}_{lane} &= w_x \mathcal{L}_{x} + w_z \mathcal{L}_{z} + w_v \mathcal{L}_{v} + w_c \mathcal{L}_{c}, \\
    \mathcal{L} &= w_s \mathcal{L}_{seg} + w_p \mathcal{L}_{plane} + w_l \mathcal{L}_{lane},
\end{aligned}
\nonumber
\setlength{\belowdisplayskip}{6pt}
\end{equation}
where $w_{\lbrack \ast \rbrack}$ represent different loss weights: $w_s$=5.0, $w_x$=2.0, $w_z$=10.0, $w_c$=10.0, and the rest are set to $1.0$. 
We use L1 loss for $\mathcal{L}_{x}, \mathcal{L}_{z}$ and use BCE loss for $\mathcal{L}_{v}$. For classification, we adopt focal loss~\cite{lin2017focal} with $\gamma$=2.0 and $\alpha$=0.25, same as~\cite{li2022bevformer,liu2022petr}.

\vspace{-1mm}
\section{Experiments}

We evaluate our method on three 3D lane benchmarks: OpenLane~\cite{chen2022persformer}, Apollo~\cite{guo2020gen} and ONCE-3DLanes~\cite{yan2022once}. 

\begin{figure*}[h]
    \centering
    \includegraphics[width=1.0\linewidth]{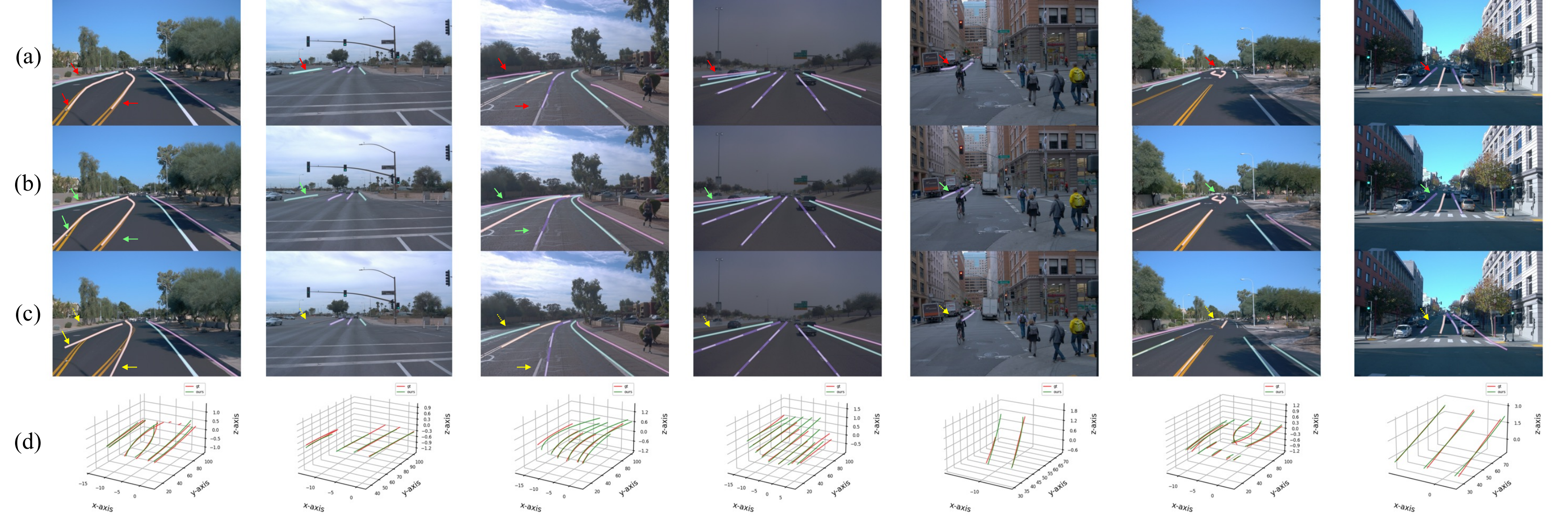}
    \caption{\textbf{Qualitative evaluation on OpenLane \textit{val} set.} The rows \textcolor[rgb]{0.8,0.15,0.15}{(a)}, \textcolor{Green}{(b)}, \textcolor[rgb]{0.91,0.69,0.04}{(c)} illustrate \textcolor[rgb]{0.8,0.15,0.15}{ground truth} 3D lanes, prediction from\textcolor{Green}{~\modelname}~and \textcolor[rgb]{0.91,0.69,0.04}{Persformer}~\cite{chen2022persformer} with 2D projection, respectively. Here, different colors indicate specific categories. Row (d) demonstrates the ground truth (\textcolor[rgb]{0.8,0.15,0.15}{red}) and prediction of \modelname\,(\textcolor{Green}{green}) in 3D space. Best viewed in color (zoom in for details). }
    \label{fig:qual_results}
    \vspace{-2mm}
\end{figure*}

\subsection{Datasets and Metrics}
\noindent\textbf{OpenLane}~\cite{chen2022persformer} is a comprehensive, large-scale benchmark for 3D lane detection, built on the Waymo dataset~\cite{sun2020scalability}. This dataset consists of 1000 segments, including 200K frames captured under various weather, terrain, and brightness conditions at a resolution of 1280$\times$1920.
OpenLane contains 880K lane annotations, which are spread across a total of 14 categories, providing a realistic and diverse set of challenges for 3D lane detection algorithms.

\noindent\textbf{Apollo Synthetic} ~\cite{guo2020gen} is generated using a game engine. It comprises over 10k images covering three distinct scenes: 1) \textit{Balanced scenes}, 2) \textit{Rarely observed scenes}, and 3) \textit{Scenes with visual variations}. The dataset includes diversified terrain structures such as highways, urban, and residential areas, as well as various illumination conditions. 

\noindent \textbf{ONCE-3DLanes}~\cite{yan2022once} is a real-world 3D lane dataset, built on the ONCE~\cite{mao2021one} dataset. It contains 211K images comprising different locations, lighting conditions, weather conditions and slop scenes. Notably, camera extrinsics are not accessible in ONCE-3DLanes. 

\noindent\textbf{Evaluation Metrics.} We follow official evaluation metrics to investigate our model on the above two datasets. 
The evaluation is formulated as a matching problem based on minimum-cost flow, where 
the lane matching cost is obtained by taking the square root of the squared sum of point-wise distances over predefined $y$s. A lane prediction is considered matched if at least 75\% of its points have a point-wise distance less than a predefined threshold of 1.5 meters~\cite{guo2020gen, chen2022persformer}. 
Errors are measured in the near range [0, 40]m and far range [40, 100]m along the heading direction. 
Moreover, the Average Precision (AP) metric is used to evaluate the performance on the Apollo Synthetic dataset~\cite{guo2020gen}.

\subsection{Experimental Settings}

\noindent\textbf{Implementation Details.}
We use an input shape of $720\times 960$ and adopt ResNet-50~\cite{he2016deep} as our backbone to extract feature maps from three scales with spatial reduction ratios of $[\frac{1}{8}$, $\frac{1}{16}$, $\frac{1}{32}]$. 
After that, we exploit FPN~\cite{lin2017feature} to generate multi-scale features, constructing a four-layer feature pyramid.
Specifically, we apply dilation convolutions to upsample small feature maps into the largest scale, \ie, the $\frac{1}{8}$ ratio, and aggregate all features as the input of our decoder.
For the decoder, we employ deformable attention~\cite{zhu2020deformable} with 4 attention heads, 8 sample points, and 256-D embeddings. Aligning with the common setting in object detection~\cite{zhangmonodetr}, we use a six-layer decoder in~\modelname~as our default version and a two-layer design as our lite version,~\modelname-Lite.

\noindent\textbf{Training.}
All our experiments are trained with the AdamW optimizer with a weight decay of 0.01. We set the learning rate to be $2\times 10^{-4}$ and use a cosine annealing scheduler. We use batch size 32 and trained models on A100 GPUs. We train the models for 24 epochs on OpenLane and ONCE-3DLanes, and 100 epochs on Apollo dataset.

\subsection{Main Results}\label{sec:exp-main-res}

\noindent \textbf{Results on OpenLane.}
We present the main results on OpenLane \textit{val} set in~\cref{tab:openlane}. Several key trends can be observed:
\textbf{1)}~\modelname~surpasses all previous methods by substantial margins. It achieves a remarkable improvement of 11.4 in F1 score and 2.5 in category accuracy compared to the previous state-of-the-art method, Persformer~\cite{chen2022persformer}.
\textbf{2)}~\modelname~achieves significant reductions in both X and Z direction errors across both the near and far range, compared to the previous SOTA. 
Notably, the errors are reduced by 0.100m/0.066m and by 0.037m/0.037m in X and Z directions within the near/far range, respectively.
\textbf{3)}~\modelname-Lite, albeit a lightweight version of~\modelname~with only two decoder layers, attains comparable results to~\modelname.
\textbf{4)} Although Persformer (Persformer-Res50) shows an increase in F1 from 50.5 to 53.0 by utilizing the same backbone and input shape as~\modelname, it still significantly lags behind~\modelname.%

\begin{table*}[t]
\begin{center}
\small
\resizebox{1.0\linewidth}{!}{
  \footnotesize
  \begin{tabular}{c|clcllll}
  \toprule
  \multicolumn{1}{c}{} &\multicolumn{1}{c}{} & \multicolumn{1}{c}{} & \multicolumn{1}{c}{} & \multicolumn{2}{c}{{X error (m)} $\downarrow$} & \multicolumn{2}{c}{{Z error (m)} $\downarrow$} \\
  \cmidrule{5-6} \cmidrule{7-8} 
  \multicolumn{1}{c}{\multirow{-2}{*}{{Scene}}} &
  \multicolumn{1}{c}{\multirow{-2}{*}{{Methods}}} & \multicolumn{1}{l}{\multirow{-2}{*}{{F1} $\uparrow$}} & \multicolumn{1}{l}{\multirow{-2}{*}{{AP} $\uparrow$}} & \multicolumn{1}{l}{\textit{near}} & \multicolumn{1}{l}{\,\textit{far}} & \multicolumn{1}{l}{\textit{near}} & \multicolumn{1}{l}{\,\textit{far}} \\ \hline\hline

\multirow{8}{*}{Balanced Scene} & 
\multicolumn{1}{l}{3DLaneNet~\cite{garnett20193d}} & 86.4 & 89.3 \quad\quad\quad & 0.068 & 0.477 & 0.015 & \textbf{0.202} \\
& \multicolumn{1}{l}{Gen-LaneNet~\cite{guo2020gen}} & 88.1 & 90.1 \quad\quad\quad &  0.061 & 0.496 & 0.012 & 0.214 \\
& \multicolumn{1}{l}{CLGo~\cite{liu2022learning}} & 91.9 & 94.2 \quad\quad\quad &  0.061 & 0.361 & 0.029 & 0.250 \\
& \multicolumn{1}{l}{PersFormer~\cite{chen2022persformer}} & 92.9 & - \quad\quad\quad &  0.054 & 0.356 & 0.010 & 0.234 \\
& \multicolumn{1}{l}{GP~\cite{li2022reconstruct}} & 91.9 & 93.8 \quad\quad\quad & \underline{0.049} & 0.387 & \underline{0.008} & \underline{0.213} \\
& \multicolumn{1}{l}{CurveFormer~\cite{bai2022curveformer}} & \underline{95.8} & \underline{97.3} \quad\quad\quad & 0.078 & \underline{0.326} & 0.018 & 0.219 \\
\cline{2-8}
& \multicolumn{1}{l}{\modelname-Lite} & 96.5 & 97.8 \quad\quad\quad & 0.035 & 0.283 & 0.012 & 0.209\\
& \multicolumn{1}{l}{\cellcolor{mgray}\modelname} & 
\cellcolor{mgray}\textbf{96.8} \footnotesize\color{RoyalBlue}{$\uparrow$\,1.0} & \cellcolor{mgray}\textbf{97.9} \footnotesize\color{RoyalBlue}{$\uparrow$\,0.6} & \cellcolor{mgray}\textbf{0.022} \footnotesize\color{RoyalBlue}{$\downarrow$\,0.027} & \cellcolor{mgray}\textbf{0.253} \footnotesize\color{RoyalBlue}{$\downarrow$\,0.073} & \cellcolor{mgray}\textbf{0.007} \footnotesize\color{RoyalBlue}{$\downarrow$\,0.001} & \cellcolor{mgray}\textbf{0.202} \footnotesize\color{RoyalBlue}{$\downarrow$\,0.011}\\

\Xhline{1pt}
\multirow{8}{*}{Rare Subset} & 
\multicolumn{1}{l}{3DLaneNet~\cite{garnett20193d}} & 74.6 & 72.0 \quad\quad\quad & 0.166 & 0.855 & 0.039 & \textbf{0.521} \\
& \multicolumn{1}{l}{Gen-LaneNet~\cite{guo2020gen}} & 78.0 & 79.0 \quad\quad\quad &  0.139 & 0.903 & 0.030 & 0.539 \\
& \multicolumn{1}{l}{CLGo~\cite{liu2022learning}} & 86.1 & 88.3 \quad\quad\quad &  0.147 & \underline{0.735} & 0.071 & 0.609 \\
& \multicolumn{1}{l}{PersFormer~\cite{chen2022persformer}} & 87.5 & - \quad\quad\quad & \underline{0.107} & 0.782 & 0.024 & 0.602 \\
& \multicolumn{1}{l}{GP~\cite{li2022reconstruct}} & 83.7 & 85.2 \quad\quad\quad & 0.126 & 0.903 & \underline{0.023} & 0.625 \\
& \multicolumn{1}{l}{CurveFormer~\cite{bai2022curveformer}} & \underline{95.6} & \underline{97.1} \quad\quad\quad & 0.182 & 0.737 & 0.039 & 0.561 \\
\cline{2-8}
& \multicolumn{1}{l}{\modelname-Lite} & 95.8 & 97.2 \quad\quad\quad  & 0.060 & 0.618 & 0.020 & 0.538 \\
& \multicolumn{1}{l}{\cellcolor{mgray}\modelname} & 
\cellcolor{mgray}\textbf{96.1} \footnotesize\color{RoyalBlue}{$\uparrow$\,0.5} & \cellcolor{mgray}\textbf{97.3} \footnotesize\color{RoyalBlue}{$\uparrow$\,0.2} & \cellcolor{mgray}\textbf{0.050} \footnotesize\color{RoyalBlue}{$\downarrow$\,0.057} & \cellcolor{mgray}\textbf{0.600} \footnotesize\color{RoyalBlue}{$\downarrow$\,0.135} & \cellcolor{mgray}\textbf{0.015} \footnotesize\color{RoyalBlue}{$\downarrow$\,0.008} & \cellcolor{mgray}\underline{0.532} \footnotesize\color{red}{$\uparrow$\,0.011} \\

\Xhline{1pt}
\multirow{8}{*}{Visual Variations} & 
\multicolumn{1}{l}{3DLaneNet~\cite{garnett20193d}} & 74.9 & 72.5 \quad\quad\quad & 0.115 & 0.601 & 0.032 & \underline{0.230} \\
& \multicolumn{1}{l}{Gen-LaneNet~\cite{guo2020gen}} & 85.3 & 87.2 \quad\quad\quad & 0.074 & 0.538 & 0.015 & 0.232 \\
& \multicolumn{1}{l}{CLGo~\cite{liu2022learning}} & 87.3 & 89.2 \quad\quad\quad & 0.084 & 0.464 & 0.045 & 0.312 \\
& \multicolumn{1}{l}{PersFormer~\cite{chen2022persformer}} & 89.6 \quad\quad\quad & - \quad\quad\quad& 0.074 & 0.430 & 0.015 & 0.266 \\
& \multicolumn{1}{l}{GP~\cite{li2022reconstruct}} & 89.9 & 92.1 \quad\quad\quad & \underline{0.060} & 0.446 & \textbf{0.011 }& 0.235 \\
& \multicolumn{1}{l}{CurveFormer~\cite{bai2022curveformer}} & \underline{90.8} & \underline{93.0} \quad\quad\quad & 0.125 & \underline{0.410} & 0.028 & 0.254 \\
\cline{2-8}
& \multicolumn{1}{l}{\modelname-Lite} & 94.0 & 95.6 \quad\quad\quad & 0.048 & 0.352 & 0.018 & 0.231 \\
& \multicolumn{1}{l}{\cellcolor{mgray}\modelname} & 
\cellcolor{mgray}\textbf{95.1} \footnotesize\color{RoyalBlue}{$\uparrow$\,4.3} & \cellcolor{mgray}\textbf{96.6} \footnotesize\color{RoyalBlue}{$\uparrow$\,3.6} & \cellcolor{mgray}\textbf{0.045} \footnotesize\color{RoyalBlue}{$\downarrow$\,0.015} & \cellcolor{mgray}\textbf{0.315} \footnotesize\color{RoyalBlue}{$\downarrow$\,0.095} & \cellcolor{mgray}\underline{0.016} \footnotesize\color{red}{$\uparrow$\,0.005} & \cellcolor{mgray}\textbf{0.228} \footnotesize\color{RoyalBlue}{$\downarrow$\,0.002} \\

\bottomrule     
\end{tabular}}

\end{center}
\vspace{-1.5mm}
\caption{\textbf{Comparison with other 3D lane detection methods on Apollo 3D Synthetic dataset with three different scenes.} \modelname~achieves the best performance over all metrics and across three scenes with notable margins. Moreover, we implement a lite version, dubbed \modelname-Lite, which consists of two decoder layers and achieves comparable results as \modelname. \textcolor{RoyalBlue}{Blue} arrows denote gains, while \textcolor{red}{red} ones denote deterioration.}
\label{tab:sota-apollo}
\vspace{-3mm}
\end{table*}

\noindent \textbf{Different Scenarios in OpenLane.} Apart from the main results on OpenLane, we also conducted comprehensive experiments across various scenarios within OpenLane \textit{val} set.
As shown in~\cref{tab:openlane-cases}, our proposed~\modelname~outperforms state-of-the-art methods by significant margins across six challenging scenarios. 
Specifically, we observed that our model performs more accurate lane detection under complex scenarios involving \textit{curves} (+11.6 \emph{w.r.t.} F1) and \textit{merges/splits} (+10.8 \emph{w.r.t.} F1), which benefits from our hybrid lane query embedding design. Additionally, our model improves significantly in the \textit{up/down} scenario (+10.0 \emph{w.r.t.} F1), suggesting that our dynamic 3D ground design enables the model to perceive the road better. These findings demonstrate the effectiveness of our proposed approach in handling diverse driving scenarios. In~\cref{fig:qual_results}, we present a qualitative comparison of~\modelname~and Persformer~\cite{chen2022persformer}, where our method performs more accurate predictions in several challenging scenarios.

However, we also observed that relatively modest improvements (+7.4/+6.3) are achieved under challenging lighting conditions, \textit{extreme weather} and \textit{night}, compared to other scenarios.
This may be attributed to the inherent constraints of the vision-centric method, which heavily relies on visual cues for perception.

\noindent \textbf{Results on Apollo.}~\cref{tab:sota-apollo} summarizes the results of our experiments on the Apollo dataset. We evaluate our method on three different scenes and study the F1 score, AP and errors, following the literature~\cite{guo2020gen}. Our~\modelname~demonstrates superiority over all scenes and metrics,
despite the performance being close to saturation.
Notably, our design significantly boosts the performance by 4.3 points in F1 and 3.6 points in AP under the visual variations scenario, which suggests the effectiveness of our dynamic ground design. Moreover, we observe that our model achieves comparable results with both two-layer and six-layer decoders.

\begin{table}[!hbp]\centering
\resizebox{\linewidth}{!}{
\begin{tabular}{c|cccc}
\toprule[1.2pt]
\textbf{Method} & \textbf{F1(\%)$\uparrow$} & \textbf{P(\%)$\uparrow$ }& \textbf{R(\%)$\uparrow$} & \textbf{CD Error(m)$\downarrow$} \\ 
\hline
3D-LaneNet~\cite{garnett20193d} & 44.73 & 61.46 & 35.16 & 0.127 \\
Gen-LaneNet~\cite{guo2020gen} & 45.59 & 63.95 & 35.42 & 0.121\\
SALAD~\cite{yan2022once} & 64.07 & 75.90 & 55.42 & 0.098 \\
PersFormer~\cite{chen2022persformer} & 74.33 & 80.30 & 69.18 & 0.074\\
\rowcolor{mgray}
LATR & 80.59 & 86.12 & 75.73 & 0.052 \\
\hline
\textit{improvement} & \color{RoyalBlue}{$\uparrow$} 6.26 & \color{RoyalBlue}{$\uparrow$ 5.82} & \color{RoyalBlue}{$\uparrow$ 6.55} & \color{RoyalBlue}{$\downarrow$ 0.022} \\
\toprule[1.2pt]  
\end{tabular}}
\caption{Comparison with state-of-the-art methods on ONCE-3DLanes \textit{val} set. 
``P'' and ``R'' denote precision and recall, respectively. 
\textcolor{RoyalBlue}{Blue} numbers indicate improvements.}
\label{tab:once}
\end{table}

\noindent \textbf{Results on ONCE-3DLanes.} We present the experimental results on ONCE-3DLanes~\cite{yan2022once} dataset in~\cref{tab:once}, where we employ a similar camera setting as~\cite{chen2022persformer}.
Despite inaccurate camera parameters, our method outperforms existing benchmarks notably on this dataset.
Compared with PersFormer,~\modelname~obtains a higher F1 score (\textbf{+6.26}) and reduces CD error by 0.022m, under a criterion of $\tau_{CD}=0.3$~\cite{yan2022once} and $prob=0.5$. 
This outcome exhibits the efficacy of our~\modelname~over different datasets.

Overall, these experiments on both realistic and synthetic datasets demonstrate the generalization and robustness of our proposed method, indicating its potential to be applied in real-world scenarios.

\subsection{Model Analysis}
We conduct a thorough analysis to validate the effectiveness of our design choices in~\modelname~ using the OpenLane-300 dataset, following the literature~\cite{chen2022persformer}. Further details concerning model complexity, and extra ablation studies (including decoder layers and input size variations) are provided in our Appendix.

\noindent\textbf{Module Ablations.}
The first and third rows in~\cref{tab:ablation_components} show the results of using learnable weights to replace the lane-level embedding. Using the learnable embedding, without prior knowledge of image features, performs much worse than its lane-aware counterpart (61.5~\vs70.4 and 45.5~\vs67.9 in F1 score). Furthermore, no matter how the lane-level embedding is obtained, noticeable performance drops are observed when the dynamic 3D ground PE is dropped from~\modelname. The gap becomes even more significant when learnable weights replace the lane-level embedding (61.5$\rightarrow$45.5). Overall, when singly applied, each component in~\cref{tab:ablation_components} significantly improves the performance over the baseline. Employing both designs further boost performance and achieved the best results. 

\begin{table}[th]
\centering
\setlength{\tabcolsep}{6pt}
\vspace{-2mm}
\resizebox{0.48\textwidth}{!}{
\begin{tabular}{cc|c|c|c}
\toprule[1.2pt]
Lane & Ground  & \multirow{2}{*}{F1 / \textit{C.Acc.}} & {X error (m)} & {Z error (m)} \\ %
Embed & Embed & & \multicolumn{1}{c|}{\makebox[0pt][c]{\hspace{-1.7em}\textit{near}\hspace{1em}}$\vert$\makebox[0pt][c]{\hspace{3.2em}\textit{far}\hspace{0.5em}}} & \multicolumn{1}{c}{\makebox[0pt][c]{\hspace{-1.7em}\textit{near}\hspace{1em}}$\vert$\makebox[0pt][c]{\hspace{3.2em}\textit{far}\hspace{0.5em}}}\\ %
\hline\hline
&  & 45.5 / 78.6 & 0.644 $\vert$ 0.491 & 0.109 $\vert$ 0.147 \\ %
\checkmark &  & 67.9 / 90.3 & 0.281 $\vert$ 0.349 & 0.102 $\vert$ 0.139 \\  %
& \checkmark & 61.5 / 87.7 & 0.352 $\vert$ 0.381 & 0.101 $\vert$ 0.140 \\ %
\checkmark & \checkmark & \textbf{70.4}  / \textbf{92.9} & \textbf{0.241} $\vert$ \textbf{0.321} & \textbf{0.097} $\vert$ \textbf{0.132} \\ %
\toprule[1.2pt]
\end {tabular}
}
\caption{\textbf{Ablation studies on designed modules.} 
``Lane Embed'' and ``Ground Embed'' denote lane-level embedding in Lane-aware Query Generator and Dynamic 3D ground PE Generator, respectively. \textit{C.Acc.} means category accuracy.
}
\vspace{-2.5mm}
\label{tab:ablation_components}
\end{table}

\noindent\textbf{Point-level Query vs. Lane-level Query.}
In~\modelname~, each query is enhanced with the point-level embedding, corresponding to a single point in the final lane prediction. Every $M$ points that belong to the same lane-level embedding are grouped together to form a complete lane.
In this part, we explore another choice, using lane-level embedding $\mathbf{Q}_{lane} \in \mathbb{R}^{N \times C}$ as the final query embeddings.
Unlike the point-level query approach, this setup requires predicting $M$ different points for each lane query.
As shown in~\cref{tab:ablation_lane_rep}, using pure lane-level queries incurs a noticeable performance drop (\eg, 70.4$\rightarrow$66.5 in terms of F1) when compared to our proposed setup.

\begin{table}[h!]
\begin{center}
\vspace{-2mm}
\setlength{\tabcolsep}{6pt}
\scalebox{0.9}{
\begin{tabular}{l|c|c|c}
\toprule[1.2pt]
 \multirow{2}{*}{Model}  & \multirow{2}{*}{F1 / \textit{C.Acc.}} & {X error (m)} & {Z error (m)} \\
& & \multicolumn{1}{c|}{\makebox[0pt][c]{\hspace{-1.7em}\textit{near}\hspace{1em}}$\vert$\makebox[0pt][c]{\hspace{3.2em}\textit{far}\hspace{0.5em}}} & \multicolumn{1}{c}{\makebox[0pt][c]{\hspace{-1.7em}\textit{near}\hspace{1em}}$\vert$\makebox[0pt][c]{\hspace{3.2em}\textit{far}\hspace{0.5em}}}\\
\hline\hline
 \multicolumn{1}{l|}{Lane only} & 66.5 / 91.3 & 0.278 $\vert$ 0.337 & 0.100 $\vert$ 0.138 \\ 
 \multicolumn{1}{l|}{Lane + Point}       & \textbf{70.4} / \textbf{92.9} & \textbf{0.241} $\vert$ \textbf{0.321} & \textbf{0.097} $\vert$ \textbf{0.132} \\
\toprule[1.2pt]
\end {tabular}
}
\caption{Different design choices for lane-aware queries.}
\vspace{-5.5mm}
\label{tab:ablation_lane_rep}
\end{center}
\end{table}

\noindent\textbf{Effect of Dynamic 3D Ground Positional Embedding.}
To evaluate the efficacy of our dynamic 3D ground positional embedding, we compared it with several alternatives.
Specifically, we explored the assignment of 3D positions to image pixels using a fixed frustum~\cite{liu2022petr} and a fixed ground plane, 
alongside our proposed method that employs an iteratively updated ground plane. 
As shown in~\cref{tab:ablation_keypos}, 
the incorporation of 3D positional information into image pixels yielded performance improvements across all evaluated methods.
As expected, we observed that generating 3D positions via a frustum exhibits inferior results compared to using a plane. 
This is reasonable, given that lanes exist on the ground.
Consequently, the adoption of a frustum introduced a significant proportion of points in the air, where lanes are non-existent.
Besides,~\cref{tab:ablation_keypos} shows that using a dynamically updated plane is better than a fixed one, demonstrating the effectiveness of our design choice.

\begin{table}[t] \centering
\setlength{\tabcolsep}{6pt}
\scalebox{0.9}{
\begin{tabular}{c|c|c|c}
\toprule[1.2pt]
\multirow{2}{*}{Methods}  & \multirow{2}{*}{F1 / \textit{C.Acc.}} & {X error (m)} & {Z error (m)} \\
& & \multicolumn{1}{c|}{\makebox[0pt][c]{\hspace{-1.7em}\textit{near}\hspace{1em}}$\vert$\makebox[0pt][c]{\hspace{3.2em}\textit{far}\hspace{0.5em}}} & \multicolumn{1}{c}{\makebox[0pt][c]{\hspace{-1.7em}\textit{near}\hspace{1em}}$\vert$\makebox[0pt][c]{\hspace{3.2em}\textit{far}\hspace{0.5em}}}\\
\hline\hline
 \multicolumn{1}{c|}{-} &  67.9 / 90.3 & 0.281 $\vert$ 0.349 & 0.102 $\vert$ 0.139 \\
 \multicolumn{1}{c|}{Fixed Frustum} & 69.1 / 91.3 & 0.277 $\vert$ 0.343 & 0.101 $\vert$ 0.137 \\
 \multicolumn{1}{c|}{Fixed Plane} & 69.2 / 91.7 & 0.263 $\vert$ \textbf{0.321} & 0.100 $\vert$ 0.133 \\
 \multicolumn{1}{c|}{Dynamic Plane} & \textbf{70.4} / \textbf{92.9} & \textbf{0.241} $\vert$ \textbf{0.321} & \textbf{0.097} $\vert$ \textbf{0.132}\\
\toprule[1.2pt]
\end {tabular}
}
\caption{\textbf{Effect of dynamic 3D ground PE.} We compare the results of different 3D positional embedding designs.}
\vspace{-4mm}
\label{tab:ablation_keypos}
\end{table}

\section{Discussion}
While our model significantly improves performance on three public datasets, even surpasses the multi-modal method~\cite{luo2022m}, it does encounter certain failure cases.
As a vision-centric method, LATR is susceptible to the loss of critical visual cues (\emph{e.g.,} glare or blur, invisible lanes in darkness, or severe shadows), which can be observed in~\cref{tab:openlane-cases}.
Incorporating inherent rich 3D geometric information in LiDAR may offer support in such challenging situations and potentially provide a more robust solution for 3D lane detection. 
The exploration of a multi-modal method is an open and intriguing avenue for future research.

\section{Conclusion}
In this work, we propose~\modelname, a simple yet effective end-to-end framework for 3D lane detection that achieves the best performance. 
It skips the surrogate view transformation
and directly performs 3D lane detection on
front-view features. 
We propose an effective lane-aware query generator to offer query informative prior
and design a hybrid embedding to enhance query perception capability by aggregating lane-level and point-level features.
Moreover, to build the 2D-3D connection, we devise a hypothetical 3D ground to encode 3D space information into 2D features. Extensive experiments show that \modelname~achieves remarkable performance. We believe that our work can benefit the community and inspire further research.

\section{Acknowledge}
This work was supported in part by Shenzhen General Program No.JCYJ20220530143600001, by the Basic Research Project No.HZQB-KCZYZ-2021067 of Hetao Shenzhen HK S\&T Cooperation Zone, by Shenzhen-Hong Kong Joint Funding No.SGDX20211123112401002, by Shenzhen Outstanding Talents Training Fund, by Guangdong Research Project No.2017ZT07X152 and No.2019CX01X104, by the Guangdong Provincial Key Laboratory of Future Networks of Intelligence (Grant No.2022B1212010001), The Chinese University of Hong Kong, Shenzhen, by the NSFC 61931024\&81922046, by Tencent Open Fund.

{\small
\bibliographystyle{ieee_fullname}
\bibliography{main}
}

\clearpage

\end{document}